\documentclass{article}

\usepackage{spconf}

\usepackage[T1]{fontenc}
\usepackage{amsmath,amssymb}

\usepackage{graphicx}
\usepackage{booktabs,array,tabularx}
\usepackage[table]{xcolor}

\usepackage{cite}
\usepackage{hyperref}

\definecolor{visionTint}{HTML}{EAF1EC}
\definecolor{audioTint}{HTML}{F5EBE2}
\definecolor{originalTint}{HTML}{ECECEA}
\definecolor{audioGainTint}{HTML}{FFF1C7}
\definecolor{gainGreen}{HTML}{276B45}
\definecolor{ControlDeltaColumn}{HTML}{EAF3FA}

\newcommand{\gain}[1]{\textcolor{gainGreen}{\textbf{#1}}}

\newcommand{\tablefont}{%
  \fontsize{9}{11}\selectfont
}

\newcommand{\compacttable}{%
  \tablefont
  \renewcommand{\arraystretch}{0.9}%
  \setlength{\aboverulesep}{1pt}%
  \setlength{\belowrulesep}{1pt}%
}

\newcommand{\Pfirst}{P\!\to\!T^{-}}
\newcommand{\Plast}{T^{-}\!\to\!P}

\newcommand{\CI}[2]{[#1,#2]}

\newcolumntype{L}{>{\raggedright\arraybackslash}X}

\title{SAME EVIDENCE, DIFFERENT JUDGMENTS:
EVIDENCE NONCOMMUTATIVITY IN VISION/SPEECH-TEXT CONFLICTS}

\name{
Zhuoyun Li
\qquad
Boxuan Wang
\qquad
Xiaowei Huang
\qquad
Yi Dong$^{\star}$
\thanks{$^{\star}$Corresponding author: yi.dong@liverpool.ac.uk}
}

\address{
School of Computer Science and Informatics,
University of Liverpool, United Kingdom
}

\begin{document}

\ninept

\maketitle

\begin{abstract}
For multimodal large language models, when images or speech conflict with accompanying text, measured text reliance can entangle modality preference with evidence position. Earlier studies of text bias often used a fixed evidence order or moved task instructions with the evidence, leaving the contribution of order unclear. In this paper, we use a paired comparison that keeps the instructions and evidence content fixed and swaps only the positions of the two sources to quantify this potential influence. Across vision and speech models, placing an image or recording after conflicting text consistently shifts answers toward its content. We also revisit previous studies and analyze why their experimental settings can lead to misleading conclusions. These findings reveal \emph{cross-modal evidence noncommutativity}: the same evidence can lead to different judgments when its order changes, and placing perceptual evidence later can increase the model's reliance on its content.
\end{abstract}

\begin{keywords}
Multimodal large language models, evidence conflict, order sensitivity, noncommutativity
\end{keywords}

\section{Introduction}
\label{sec:introduction}

Multimodal large language model (MLLM) assistants can process charts, document pages, and speech recordings alongside user-provided descriptions and questions~\cite{qwen25vl,qwen25omni}. These words may reflect the user's interpretation or recollection of the attachment.
Spelling mistakes or memory errors can make that description conflict with the attachment. For example, a user may describe Method A as the best performer in a results plot that ranks Method B first. A recording may state that tomorrow's meeting starts at 9 a.m., while the user writes, ``Schedule tomorrow's meeting at 8 a.m. based on the manager's recording.'' The model should understand the attachment and answer or act according to the source specified by the user.


Studies of cross-modal conflict find that vision-language models often rely excessively on text when it contradicts an image~\cite{mixedsignals,wordsorvision,contextvqa,robustconflict}. Research on input order shows that rearranging long-context information, answer options, multiple images, or image--text pairs can change model responses~\cite{ordermatters,lostmiddle,optionorder}. Together, these findings raise a question: \emph{how much of the measured text reliance really reflects modality preference, while how much actually reflects the position impact of the competing evidence?}


Prior work has tried to examine how evidence order affects modality preference~\cite{wordsorvision}. However, we find that this comparison swaps the image with a text block containing the description, task instructions, and cautionary warnings. This changes both evidence order and instruction placement, while the warnings may also influence how the model treats the text. This design leaves the order's impact still entangled. Other conflict benchmarks measure modality preference under a fixed input layout~\cite{mixedsignals,mcrbench}, without enough exploration of evidence order for different modalities, which may lead to the over- or underestimation of the modality bias.

In this paper, we address the gaps by redesigning the prompt structure around everyday requests. We place the instruction to answer from the attachment before both evidence sources and manually construct faithful and conflicting textual descriptions for existing multimodal benchmarks. We then swap only the positions of the two sources, keeping all other components fixed, and compare the model’s relative reliance on the two modalities. Controls involving unrelated text, within-modality evidence swaps, and alternative instruction wording allow us to quantify \textit{cross-modal evidence noncommutativity}.

Our contributions are as follows. \textbf{(1) Cross-modal evidence noncommutativity.} We measure how judgments change when identical evidence is reordered. \textbf{(2) Evidence across vision and speech.} We demonstrate systematic noncommutativity across model families and extend the evaluation to diverse audio tasks, addressing a gap in fixed-order audio conflict benchmarks. \textbf{(3) Protocol diagnosis.} We show that moving task instructions with the evidence can confound order comparisons; our simple protocol keeps the instructions fixed to isolate the effect of evidence order.

\begin{figure}
    \centering
    \includegraphics[width=0.9\linewidth]{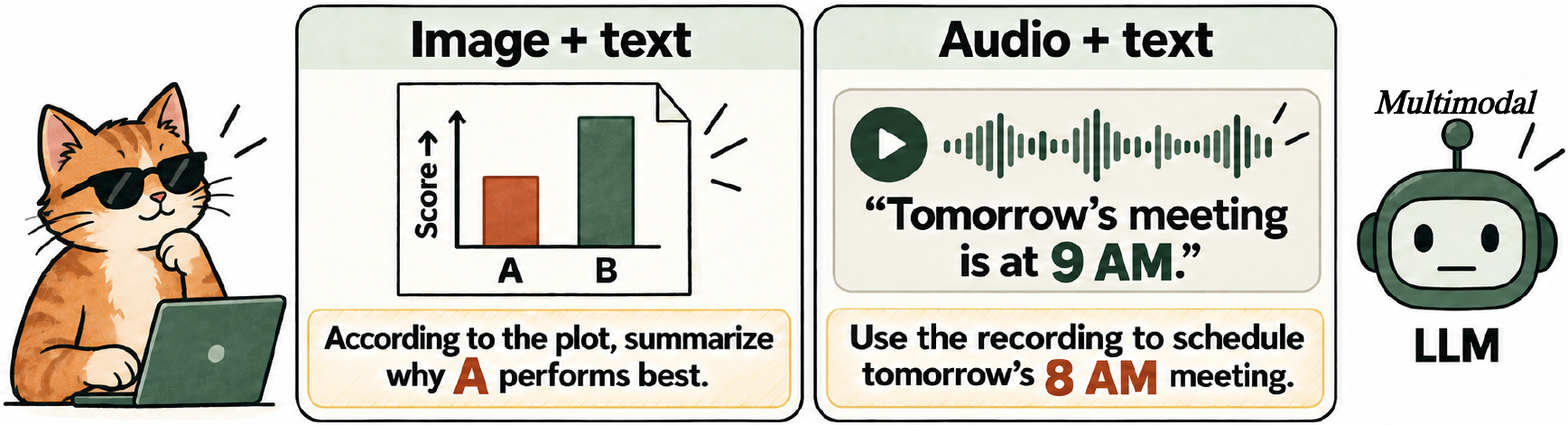}
    \caption{Everyday requests in which user descriptions conflict with image or speech attachments.}
    \label{fig:placeholder}
\end{figure}
\section{Related Work}
\label{sec:related}

\textbf{Cross-modal conflict and text reliance}
Studies of cross-modal conflict examine how models use competing sources. Mixed Signals finds that modality reliance varies with the model and task difficulty~\cite{mixedsignals}. Words or Vision examines when models follow misleading text and how image--text order affects their answers~\cite{wordsorvision}. ConText-VQA shows that persuasive false descriptions can override visual evidence~\cite{contextvqa}. Other work studies conflicts among images, text, and a model's prior knowledge~\cite{robustconflict,unraveling,insight}. Together, these studies show how textual misinformation influences vision-language models. Under a fixed template, the measured text reliance can also reflect evidence position.
The concurrent preprint~\cite{samesemantics} studies modality and order in conflicts between parametric knowledge and external knowledge presented as text or rendered images. Imitating Words or Vision, MCR-BENCH tests the textual preference but still leaves the order effect blank~\cite{mcrbench}. In this paper, we pair a natural image or real speech recording with an external description and measure what changes when the two sources exchange positions.

\noindent\textbf{Input order influence}
Language-model responses vary with information position and answer-option order~\cite{lostmiddle,optionorder}. Multimodal models are also sensitive to image and image--text arrangements~\cite{ordermatters}. We counterbalance option assignments while preserving each source's content and semantic role. Matched, conflicting, and unrelated descriptions then distinguish order effects across evidence relations.
\nocite{fragileflow,scope,azulay2026jailbreaking,pathmark,10980439,chen2026promptsleavebehavioralfingerprints,wang-etal-2026-chain,wang2026diveambiguityainspiredmultiagents,wang2025rethinkingmultiagentintelligencelens}


\section{Method}
\label{sec:method}

\subsection{Evidence and paired swaps}

We use $P$ for perceptual evidence, an image or speech recording, and $T$ for the accompanying text. Each question has an attachment-supported answer $y_P$ and a competing answer $y_T\ne y_P$. We construct three text conditions: conflicting $T^{-}$ supports $y_T$; matched $T^{+}$ agrees with $P$; and unrelated $T^{0}$ supports neither answer. The superscripts $-$, $+$, and $0$ identify these conditions throughout.
For each text condition, we keep task instruction $I$ before both sources and question/options $q$ after them. The paired inputs are
\begin{equation}
x_{P\to T}=(I,P,T,q),\qquad
x_{T\to P}=(I,T,P,q).
\label{eq:swap}
\end{equation}
Only $P$ and $T$ exchange positions; the system prompt, evidence content, and generation settings stay fixed. We evaluate both A/B label assignments and combine results by the underlying answer to counterbalance option-label bias.

\subsection{Order-sensitivity metrics}

Let $A$ denote the rate at which the model selects the attachment-supported answer $y_P$. Our primary metric is the change in this rate under conflicting text:
\begin{equation}
\Delta A^{-}
=A(T^{-}\!\to\!P)-A(P\!\to\!T^{-}).
\label{eq:delta}
\end{equation}
We report $\Delta A^{-}$ in percentage points. A positive value means the model follows $P$ more often when it appears last.

We also measure changes in answer probabilities within each conflict pair. For a model $p_\theta$, let $\ell_P,\ell_T\in\{\mathrm{A},\mathrm{B}\}$ denote the labels assigned to $y_P,y_T$ in the current option ordering. Using their probabilities at the first answer token, we define
\begin{align}
s_\theta(x)&=\log\frac{p_\theta(\ell_P\mid x)}{p_\theta(\ell_T\mid x)},
\label{eq:odds}\\
\mathcal C_\theta&=s_\theta(x_{T\to P})-s_\theta(x_{P\to T}).
\label{eq:commutator}
\end{align}
A positive log-odds swap score $\mathcal C_\theta$ indicates a shift toward $y_P$ when $P$ is placed last. We report the mean across paired evaluations.

\subsection{Controls and uncertainty}

The matched and unrelated conditions test order sensitivity when the sources agree or the text provides no evidence for either answer. Their selection-rate changes are denoted by $\Delta A^{+}$ and $\Delta A^{0}$, with the question and candidate answers held fixed across all three conditions. To compare conflict with the unrelated control, we report
\begin{equation}
D=\Delta A^{-}-\Delta A^{0}.
\label{eq:interaction}
\end{equation}
A positive $D$ indicates a larger shift toward $P$ under conflicting than unrelated text. We compute confidence intervals with a paired bootstrap over original images or recordings, keeping each example's evidence orders, option assignments, and text conditions together.

\section{Experiments}
\label{sec:experiments}

\subsection{Experimental setup}

\begin{figure}
    \centering
    \includegraphics[width=1\linewidth]{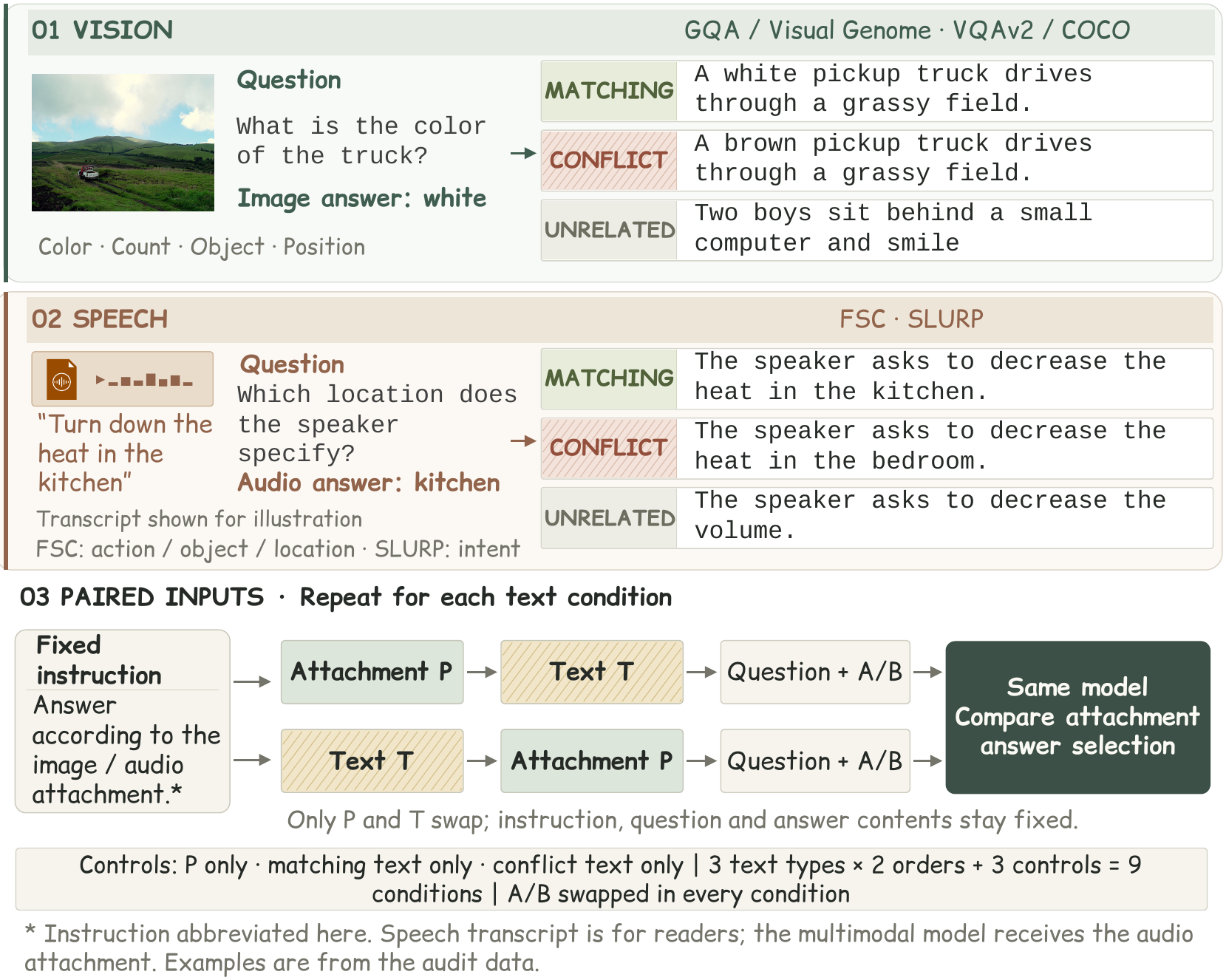}
    \vspace{-4pt}
    \caption{Controlled evidence swaps in vision--text and speech--text. Keeping instructions and content fixed lets us measure how order changes which source the model follows.}
    \label{fig:evidence-swap}
\end{figure}

\noindent\textbf{Visual data.} We manually inspect 240 natural images each from GQA/Visual Genome~\cite{gqa,visualgenome} and VQAv2/COCO~\cite{vqav2,coco}. To construct $T^{-}$, we change the queried fact (including \textit{color, count, object identity, and location}) and keep other scene details unchanged wherever possible. For example, for a cup-color question, we change the color while retaining the object and its location. Unrelated text $T^{0}$ comes from other images and supports neither candidate answer.

\noindent\textbf{Speech data.} We select 240 recordings each from Fluent Speech Commands (FSC)~\cite{fsc} and SLURP~\cite{slurp}. We allocate 80 examples to each FSC field (including  \textit{action, object, and location}) and change only one field when constructing $T^{-}$. The SLURP subset covers 15 broad intent categories, with 16 examples per category. Conflicting text $T^{-}$ comes from another intent within the same broad category, while unrelated text $T^{0}$ comes from a different category. The exploratory and confirmatory speech sets are disjoint.

\noindent\textbf{Models.} Vision models are Qwen2.5-VL-3B-Instruct~\cite{qwen25vl}, Idefics3-8B-Llama3~\cite{idefics3}, and Pixtral-12B~\cite{pixtral}. Speech models are Qwen2-Audio-7B-Instruct~\cite{qwen2audio}, Qwen2.5-Omni-7B~\cite{qwen25omni}, and Phi-4 Multimodal~\cite{phi4}. These recent models support swapping the order of perceptual and text tokens.

\noindent\textbf{Input conditions.} Each example has nine input conditions: three single-source baselines ($P$ only, $T^{+}$ only, and $T^{-}$ only) and six paired inputs (both orders of $P$ with each of $T^{+}$, $T^{-}$, and $T^{0}$). \textit{We run each condition twice}, reversing the A/B option positions to counterbalance option-letter preference~\cite{optionorder,mcselectors}.

\noindent\textbf{Prompt construction.} 
We constructed the prompts to reflect everyday requests combining an attachment, a description, and a question.
It asks the model to answer from the attachment, introduces the user's description naturally, and leaves its correctness unspecified. The visual prompt is ``Answer the question according to the image attachment, and here is also the image description.'' The speech prompt uses the corresponding audio-attachment wording.

\subsection{Quantifying cross-modal evidence noncommutativity}
\label{sec:mainresults}

\begin{table}[t]
\centering
\caption{Main image--text and speech--text results. $P$ denotes perceptual evidence; $T^{-}$ denotes conflicting text. The number is the question selection accuracy (pp). $\Delta A^{-}$ is the change of rate. $\mathcal C$ is the mean log-odds swap score with its 95\% confidence interval.}
\label{tab:main}
\compacttable
\setlength{\tabcolsep}{1.5pt}
\begin{tabular*}{\columnwidth}{@{\extracolsep{\fill}}lrrr>{\columncolor{ControlDeltaColumn}}rrl@{}}
\toprule
Model & $P$ only & $\Pfirst$ & $\Plast$ & $\Delta A^{-}$ & $\mathcal C$ [95\% CI] \\
\midrule
\rowcolor{visionTint}
\multicolumn{6}{@{}l}{\textbf{\textit{Vision: GQA / Visual Genome}}}\\
Qwen-VL & 91.5 & 28.3 & 45.0 & \textbf{+16.7} & 0.65 [0.50, 0.80] \\
Idefics3 & 89.6 & 41.0 & 67.7 & \textbf{+26.7} & 3.83 [3.37, 4.29] \\
Pixtral & 91.5 & 23.5 & 35.6 & \textbf{+12.1} & 1.38 [1.03, 1.74] \\
\midrule
\rowcolor{visionTint}
\multicolumn{6}{@{}l}{\textbf{\textit{Vision: VQAv2 / COCO}}}\\
Qwen-VL & 99.2 & 70.0 & 85.0 & \textbf{+15.0} & 1.07 [0.94, 1.21] \\
Idefics3 & 97.5 & 66.9 & 95.0 & \textbf{+28.1} & 4.77 [4.40, 5.14] \\
Pixtral & 97.1 & 61.7 & 75.8 & \textbf{+14.2} & 1.87 [1.58, 2.15] \\
\midrule
\rowcolor{audioTint}
\multicolumn{6}{@{}l}{\textbf{\textit{Speech: Fluent Speech Commands}}}\\
Qwen-A & 98.1 & 20.2 & 47.7 & \textbf{+27.5} & 3.57 [3.24, 3.91] \\
Qwen-O & 99.4 & 42.7 & 53.8 & \textbf{+11.0} & 0.42 [0.21, 0.63] \\
Phi-4 & 97.9 & 27.3 & 53.3 & \textbf{+26.0} & 1.59 [1.36, 1.82] \\
\midrule
\rowcolor{audioTint}
\multicolumn{6}{@{}l}{\textbf{\textit{Speech: SLURP}}}\\
Qwen-A & 89.2 & 6.0 & 31.3 & \textbf{+25.2} & 2.44 [2.17, 2.72] \\
Qwen-O & 91.0 & 19.4 & 29.2 & \textbf{+9.8} & 0.66 [0.51, 0.81] \\
Phi-4 & 88.3 & 15.0 & 29.2 & \textbf{+14.2} & 1.74 [1.51, 1.98] \\
\bottomrule
\end{tabular*}
\par\vspace{2pt}
\end{table}

We first test whether models that understand an attachment change their judgments when its position relative to conflicting text is reversed. As shown in Table~\ref{tab:main}, with attachments alone, accuracy ranges from 88.3--99.4\% across both vision and speech settings. These baselines indicate that the models can generally recover the relevant information from the attachments.
Adding conflicting text often draws answers away from the attachment under both input orders, consistent with earlier findings on text reliance. 

Crucially, \textbf{placing the attachment last systematically shifts judgments back toward its content}. Attachment-answer selection increases by 9.8--28.1 percentage points across all twelve model--dataset combinations. The mean log-odds swap score is also positive throughout, with every 95\% confidence interval excluding zero. This consistent shift not only demonstrates noncommutativity in the positions of perceptual and textual evidence, but also suggests better judgment performance when evidence is presented later.

\subsection{Controlling for the effect of reordering}
\label{sec:controls}

We next test whether the attachment-last advantage reflects a general effect of reordering. We repeat the swaps with matched $T^{+}$ and unrelated $T^{0}$, keeping the attachment, question, and candidate answers fixed. The unrelated condition provides a baseline for order effects when the text supports neither answer. We subtract this baseline from the conflict condition to obtain $D=\Delta A^{-}-\Delta A^{0}$.

\begin{table}[t]
\centering
\caption{Order effects under evidence control. Values show changes in attachment-answer selection when $P$ appears last (pp). $D$=$\Delta A^{-}$-$\Delta A^{0}$ shows the net influence under evidence relation control. The arrows are omitted. $P$ and $T^+$ mean only use that evidence.}
\label{tab:controls}

\compacttable
\small
\setlength{\tabcolsep}{2pt}
\begin{tabular*}{\columnwidth}{@{\extracolsep{\fill}}lrrrr>{\columncolor{ControlDeltaColumn}}rrr>{\columncolor{ControlDeltaColumn}}rr@{}}
\toprule
Model & $P$ & $T^+$ & $P T^+$ & $T^+ P$ & $\Delta A^{+}$ & $P T^0$ & $T^0 P$ & $\Delta A^{0}$ & $D$ \\
\midrule
\rowcolor{visionTint}\multicolumn{10}{l}{\textbf{\textit{Vision: GQA / Visual Genome}}} \\
Qwen-VL & 91.5 & 98.3 & 97.5 & 97.1 & -0.4 & 88.5 & 90.0 & +1.5 & \textbf{+15.2} \\
Idefics3 & 89.6 & 98.5 & 98.8 & 92.5 & -6.3 & 88.3 & 86.3 & -2.1 & \textbf{+28.8} \\
Pixtral & 91.5 & 98.3 & 99.0 & 99.2 & +0.2 & 87.9 & 90.6 & +2.7 & \textbf{+9.4} \\
\midrule
\rowcolor{visionTint}\multicolumn{10}{l}{\textbf{\textit{Vision: VQAv2 / COCO}}} \\
Qwen-VL & 99.2 & 99.8 & 100.0 & 99.8 & -0.2 & 98.8 & 99.0 & +0.2 & \textbf{+14.8} \\
Idefics3 & 97.5 & 100.0 & 99.8 & 97.5 & -2.3 & 96.5 & 96.5 & +0.0 & \textbf{+28.1} \\
Pixtral & 97.1 & 100.0 & 99.4 & 99.8 & +0.4 & 97.1 & 97.1 & +0.0 & \textbf{+14.2} \\
\midrule
\rowcolor{audioTint}\multicolumn{10}{l}{\textbf{\textit{Speech: Fluent Speech Commands}}} \\
Qwen-A & 98.1 & 97.7 & 98.3 & 99.2 & +0.8 & 89.8 & 95.6 & +5.8 & \textbf{+21.7} \\
Qwen-O & 99.4 & 100.0 & 100.0 & 100.0 & +0.0 & 94.0 & 94.6 & +0.6 & \textbf{+10.4} \\
Phi-4 & 97.9 & 99.4 & 100.0 & 100.0 & +0.0 & 89.8 & 91.9 & +2.1 & \textbf{+24.0} \\
\midrule
\rowcolor{audioTint}\multicolumn{10}{l}{\textbf{\textit{Speech: SLURP}}} \\
Qwen-A & 89.2 & 100.0 & 100.0 & 100.0 & +0.0 & 85.0 & 88.5 & +3.5 & \textbf{+21.7} \\
Qwen-O & 91.0 & 100.0 & 100.0 & 100.0 & +0.0 & 88.8 & 87.1 & -1.7 & \textbf{+11.5} \\
Phi-4 & 88.3 & 100.0 & 100.0 & 99.8 & -0.2 & 81.7 & 84.0 & +2.3 & \textbf{+11.9} \\
\bottomrule
\end{tabular*}
\end{table}

In Table~\ref{tab:controls}, $\Delta A^{+}$ shows that matched text usually produces little change. An exception is Idefics3 on GQA, where placing the image last reduces accuracy by 6.3 pp. In this setting, matched-text-only accuracy is 98.5\%, compared with 89.6\% for the image alone. One possible explanation is that placing the image last shifts reliance toward the source the model interprets less accurately.

With unrelated text, order effects $\Delta A^{0}$ are generally small and vary in sign. After subtracting this baseline, $D$ remains positive across all twelve model--dataset combinations, ranging from 9.4--28.8 pp for vision and 10.4--24.0 pp for speech. Thus, \textbf{the attachment-last advantage persists after accounting for the unrelated-text order effect}, strengthening the evidence for cross-modal noncommutativity under conflict.

\subsection{Ruling out explanations}
\label{sec:robustness}

We next investigate what might explain the consistent attachment-last advantage. Both evidence sources precede the answer position and are available during prediction, yet exchanging their positions systematically changes the model's judgments. We probe several possible contributors to this order sensitivity and rule them out.

\noindent\textbf{Text-only recency.} Removing the image and swapping correct and conflicting text yields effects from $-11.4$ to $+55.6$ pp across six combinations, with inconsistent directions. In the vision tests, all six combinations favor the image more when it appears last. Text recency remains a possible contributor to the observed effects.

\noindent\textbf{Explicit source-priority instructions.} In the vision tests, adding ``If the description conflicts with the image, prioritize the image'' remains 89--145\% of the original order effect. For FSC with Qwen2-Audio, neutral, main, and stronger speech-priority prompts yield effects of $+22.5$, $+27.5$, and $+34.6$ pp, respectively. MLLMs remain sensitive to order even when explicitly told which source to follow.

\noindent\textbf{Attachment wording.} On the same 60 exploratory SLURP examples, the original prompt yields $+21.7$ pp. Changing it to ``Answer the question according to the audio attachment.'' yields $+23.3$ pp, with a 95\% confidence interval of $\CI{14.2}{33.3}$. The paired difference is $+1.7$ pp, $\CI{-1.7}{5.0}$, and its interval includes zero. This exploratory comparison applies to these examples.

\noindent\textbf{Evidence distance.} Inserting 64 or 256 non-informative tokens between the visual and textual evidence increases the order effect in five combinations and decreases it for Qwen2.5-VL on GQA. The effect of evidence distance therefore varies across models.

\noindent\textbf{Free-form responses.} On FSC with Qwen2-Audio, the order effect remains $+33.3$ pp, $\CI{27.5}{39.2}$, showing that order also matters in free-form responses. Unrelated text can elicit a third answer, so comparisons with the conflict condition need a scoring rule that accounts for such responses.

\subsection{Revisiting existing protocols}
\label{sec:external}

The attachment-last advantage persists across the prompt and response variants tested above. We now revisit earlier studies to understand why their input designs can lead to different conclusions about modality preference.

\begin{figure}
    \centering
    \includegraphics[width=1\linewidth]{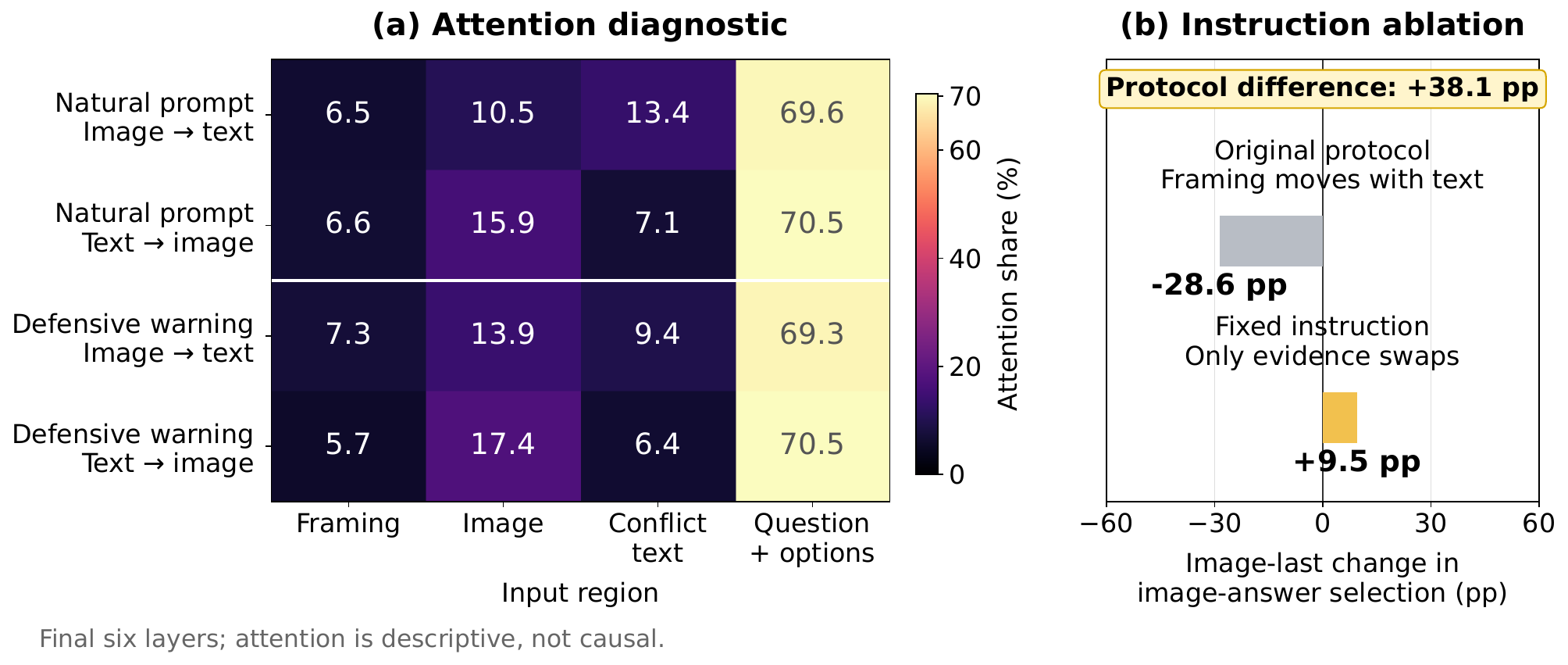}
    \vspace{-4pt}
    \caption{Exploratory Qwen2.5-VL-3B diagnostics on 20 examples: (a) attention moved by warning; (b) instruction placement ablation.}
    \label{fig:protocol-diagnostics}
\end{figure}

\noindent\textbf{Reproducing and diagnostics.} \emph{Words or Vision}~\cite{wordsorvision} reports greater text preference in Phi-3.5 when the image follows the conflicting text. We observe the same pattern with Qwen2.5-VL-3B-Instruct on 1,000 VQAv2 examples using its warning prompt and free-form protocol: moving the image last reduces image-answer accuracy from 42.6\% to 28.9\% ($-13.7$ pp).
In this protocol, the image exchanges positions with a block containing the task framing, a warning, and the conflicting description, so the task instructions move along with the text. The warning also describes the text as potentially irrelevant, incomplete, or inaccurate and advises caution. Prior studies of negative-token attention and negated prompts~\cite{negativeattention,negatedprompts} suggest that such wording may influence how the model processes the evidence. Our exploratory attention comparison shows different attention distributions under the natural and warning prompts (Fig.~\ref{fig:protocol-diagnostics}a), providing descriptive evidence of this sensitivity.

We therefore apply our protocol to the same Qwen2.5-VL model: remove the warning, fix the instruction to answer from the image before both sources, and place the question and A/B options after them. Only the image and description exchange positions. On 807 examples answered correctly from each source in isolation, moving the image last increases image-answer selection from 14.9\% to 32.6\% ($+17.7$ pp), matching the direction found in our main experiments.
To test instruction placement directly, we retain the original framing and warning and fix both before the evidence (Fig.~\ref{fig:protocol-diagnostics}b). On the same examples, moving the image last changes image-answer selection by $-28.6$ pp under the original protocol and $+9.5$ pp once the framing is fixed, a difference of $+38.1$ pp. This paired ablation shows that moving task instructions with the evidence can reverse the conclusion about which order favors image-supported answers.

\noindent\textbf{Extending the evaluation to MCR-BENCH.} MCR-BENCH~\cite{mcrbench} builds on \emph{Words or Vision} and compares faithful, adversarial, and irrelevant text under a fixed audio--text order. We evaluate both evidence orders on 1,000 examples each for audio question answering (AQA), speech emotion recognition (SER), and vocal sound classification (VSC). As shown in Table~~\ref{tab:external}, with the original warning retained, moving audio after conflicting text increases accuracy by $+17.0$, $+15.1$, and $+51.3$ pp, respectively. Under the audio-attachment prompt without the warning, the corresponding gains are $+18.2$, $+11.5$, and $+44.1$ pp. The average gain across the three tasks is $+27.8$ pp with the warning and $+24.6$ pp without it, showing that placing audio last improves accuracy under both prompts.
The comparisons show why both evidence orders matter when assessing modality preference. Instruction placement can reverse which order favors image-supported answers, while a fixed audio--text layout leaves the gains from placing audio last unmeasured.

\section{Discussion}
\label{sec:discussion}

\textbf{Order contributes to measured text reliance}
Text reliance under a fixed template reflects modality, position, distance from the question, and conflict between sources. In our tests, moving the same image or speech recording later shifts answers toward it, while average modality preferences differ across models. Paired swaps measure how order contributes to source reliance alongside possible textual shortcuts.
A source's influence depends on where and how it appears in the input, including whether another source contradicts it. Evaluations should test both $P$$\to$$T$ and $T$$\to$$P$ and report average modality reliance alongside the change caused by swapping order. 

\noindent\textbf{Implications for deployed systems}
Speech assistants often combine audio with an ASR transcript, a user's description, device state, or retrieved information. Our results suggest that the placement of these sources should be considered alongside their origin and consistency. The same issue arises in multimodal retrieval-augmented generation and agents, where retrieval ranking, attachment placement, and tool-return order determine the input sequence. Reordering these sources may change the answer even when their content stays the same.
A practical check is to reorder key evidence and query the model again. If the answer changes, the system can review the sources, flag the disagreement, or seek clarification before a high-risk action. 

\begin{table}[t]
\centering
\caption{External benchmarks under original and aligned protocols in pp. $P$ denotes images for Words or Vision (WoV) \cite{wordsorvision} and audio for MCR-BENCH (MCR) \cite{mcrbench}. Yellow highlights gains from placing audio last with the original warning retained.}
\label{tab:external}
\compacttable
\setlength{\tabcolsep}{1.5pt}
\begin{tabular*}{\columnwidth}{@{\extracolsep{\fill}}lrrrrrr@{}}
\toprule
 & \multicolumn{3}{c}{Original warning} & \multicolumn{3}{c}{Aligned protocol} \\
\cmidrule(lr){2-4}\cmidrule(l){5-7}
Task & $\Pfirst$ & $\Plast$ & $\Delta A^{-}$ & $\Pfirst$ & $\Plast$ & $\Delta A^{-}$ \\
\midrule
WoV / VQAv2 & \cellcolor{originalTint}42.6 & \cellcolor{originalTint}28.9 & \cellcolor{originalTint}\textbf{$-13.7$} & 14.9 & 32.6 & \gain{+17.7} \\
MCR / AQA & \cellcolor{originalTint}1.4 & \cellcolor{audioGainTint}18.4 & \cellcolor{audioGainTint}\gain{+17.0} & 2.4 & 20.6 & \gain{+18.2} \\
MCR / SER & \cellcolor{originalTint}0.0 & \cellcolor{audioGainTint}15.1 & \cellcolor{audioGainTint}\gain{+15.1} & 0.1 & 11.6 & \gain{+11.5} \\
MCR / VSC & \cellcolor{originalTint}12.7 & \cellcolor{audioGainTint}64.0 & \cellcolor{audioGainTint}\gain{+51.3} & 20.9 & 65.0 & \gain{+44.1} \\
\bottomrule
\end{tabular*}
\par\vspace{2pt}
\end{table}

\section{Conclusion}
\label{sec:conclusion}

In this paper, we demonstrate cross-modal evidence noncommutativity: the same evidence can lead to different judgments when its order changes. Across vision and speech models, placing an attachment after conflicting text consistently shifts judgments toward its content. Revisiting earlier evaluations also reveals protocol confounds that can reverse the measured direction of these shifts and lead to misleading conclusions about modality preference.

Our additional experiments probe several possible explanations and show that the shifts persist under alternative instructions and response formats. Understanding their internal causes will require further analysis of perceptual representations, positional encoding, attention, and decoding.
We are also extending this study to multiple perceptual sources $P_1,\ldots,P_m$ and textual sources $T_1,\ldots,T_n$ to examine how different source combinations and orderings affect the interactions between evidence sources and the resulting judgments.

\clearpage

\begin{center}
\bfseries ACKNOWLEDGEMENTS
\end{center}
This work is partially funded by the European Union (under grant agreement ID 101212818). Views and opinions expressed are however those of the author(s) only and do not necessarily reflect those of the European Union or European Health and Digital Executive Agency (HADEA). Neither the European Union nor the granting authority can be held responsible for them. 
This work is partially supported by Innovate UK through AI-PASSPORT under Grant 10126404. This work was awarded a grant by the AI Security Institute (AISI) via the Alignment Project
(Rare-Event Estimation in Large Language Models via Subset Simulation) and funded by EPSRC.
Yi's contribution is partially supported through the Royal Society international exchanges programme and in part by the Engineering and Physical Sciences Research Council, through funding from RAi UK [EP/Y009800/1].

The authors declare no relevant financial or non-financial conflicts of interest.

\begin{center}
\bfseries COMPLIANCE WITH ETHICAL STANDARDS
\end{center}
No new human participants were recruited and no new human-subject data were collected for this study. All experiments use existing publicly available benchmark datasets under their respective terms of use.

\bibliographystyle{IEEEbib}
\bibliography{references}

\end{document}